\documentclass[preprint,12pt]{elsarticle}

\usepackage{amssymb}
\usepackage{graphicx}
\usepackage{multirow}
\usepackage{xcolor,color,stfloats}
\usepackage{subfigure}
\usepackage{amsmath}

\usepackage{float}
\usepackage{booktabs}
\usepackage{lineno}

\journal{Pattern Recognition}

\begin{document}

\begin{frontmatter}

%% Title, authors and addresses

%% use the tnoteref command within \title for footnotes;
%% use the tnotetext command for the associated footnote;
%% use the fnref command within \author or \address for footnotes;
%% use the fntext command for the associated footnote;
%% use the corref command within \author for corresponding author footnotes;
%% use the cortext command for the associated footnote;
%% use the ead command for the email address,
%% and the form \ead[url] for the home page:
%%
%% \title{Title\tnoteref{label1}}
%% \tnotetext[label1]{}
%% \author{Name\corref{cor1}\fnref{label2}}
%% \ead{email address}
%% \ead[url]{home page}
%% \fntext[label2]{}
%% \cortext[cor1]{}
%% \address{Address\fnref{label3}}
%% \fntext[label3]{}

\title{Learning Scene-specific Object Detectors Based on a Generative-Discriminative Model with Minimal Supervision}

%% use optional labels to link authors explicitly to addresses:
%% \author[label1,label2]{<author name>}
%% \address[label1]{<address>}
%% \address[label2]{<address>}

\author[cugm]{Dapeng Luo}
\author[cugm]{Zhipeng Zeng}
\author[hust]{Nong Sang}
\author[nan]{Xiang Wu}
\author[cuga]{Longsheng Wei\corref{cor1}}
\author[cugm]{Quanzheng Mou}
\author[cuga]{Jun Cheng}
\author[hui]{Chen Luo}

\cortext[cor1]{Corresponding author}
\address[cugm]{School of Mechanical Engineering and Electronic Information, China University of Geosciences, Wuhan 430074 China}
\address[hust]{School of Automation, Huazhong University of Science and Technology, Wuhan 430074 China}
\address[cuga]{School of Automation, China University of Geosciences, Wuhan 430074 China}
\address[nan]{School of Automation, Nanjing University of Science and Technology, Nangjing  210094 China}
\address[hui]{Huizhou School Affiliated to Beijing Normal University, Huizhou 516002, China}

\begin{abstract}
%% Text of abstract
One object class may show large variations due to diverse illuminations, backgrounds and camera viewpoints. Traditional object detection methods often perform worse under unconstrained video environments. To address this problem, many modern approaches model deep hierarchical appearance representations for object detection. Most of these methods require a time-consuming training process on large manual labelling sample set. In this paper, the proposed framework takes a remarkably different direction to resolve the multi-scene detection problem in a bottom-up fashion. First, a scene-specific objector is obtained from a fully autonomous learning process triggered by marking several bounding boxes around the object in the first video frame via a mouse. Here the human labeled training data or a generic detector are not needed. Second, this learning process is conveniently replicated many times in different surveillance scenes and results in particular detectors under various camera viewpoints. Thus, the proposed framework can be employed in multi-scene object detection applications with minimal supervision. Obviously, the initial scene-specific detector, initialized by several bounding boxes, exhibits poor detection performance and is difficult to improve with traditional online learning algorithm. Consequently, we propose Generative-Discriminative model to partition detection response space and assign each partition an individual descriptor that progressively achieves high classification accuracy. A novel online gradual optimized process is proposed to optimize the Generative-Discriminative model and focus on the hard samples: the most informative samples lying around the decision boundary. The output is a hybrid classifier based scene-specific detector which achieves decent performance under different scenes. Experimental results on six video datasets show our approach achieves comparable performance to robust supervised methods, and outperforms the state of the art self-learning methods under varying imaging conditions.

\end{abstract}

\begin{keyword}
Object detection \sep Unsupervised learning, Generative-Discriminative model
%% keywords here, in the form: keyword \sep keyword

%% MSC codes here, in the form: \MSC code \sep code
%% or \MSC[2008] code \sep code (2000 is the default)

\end{keyword}

\end{frontmatter}

%%
%% Start line numbering here if you want
%%
%%\linenumbers

%% main text
\section{Introduction}
\label{S:i}

With the development of intelligent surveillance systems, pedestrian and vehicle detection approaches have garnered profound interest from engineers and scholars. Many impressive works \cite{viola2001rapid,dalal2005histograms,wang2009hog,felzenszwalb2010object,li2013component,dollar2014fast,Redmon2016YOLO9000} have been published in the last several years. However, object detection and recognition remain a considerably difficult issue in highly populated public places such as airports, train stations and urban arterial roads, which have distributed multiple surveillance cameras with different viewpoints, illuminations and backgrounds in cluttered environments. One object class may show large inter-class variations under these different imaging conditions. A substantial amount of training data need to be collected and labeled to model one object category detector based on statistical learning. Without this, the detector, trained in constrained video environments, will deliver poor performance under different environmental conditions. How to robustly and stably locate objects in arbitrary video environments with minimal supervision is still an open issue.

Transfer learning methods \cite{Wang2014Scene,pang2011transferring} can be used to learn different view-point scene detectors from pre-training models, which reduce the efforts involved in collecting samples and retraining in response to such variations in appearance. However, negative transfer often occurs when transfering very different scenes \cite{rosenstein2005transfer}, significantly influencing the performance of target scene detectors and limiting the application of the transfer learning.

The Multi-view object detection method is an alternative strategy to locate the object in various view-points, which minimizes the influence of changing imaging conditions. However, these approaches increase the discriminability of detectors through relatively complex additional stages involving view-invariant features \cite{torralba2007sharing,leibe2008robust,uijlings2013selective,ko2015view}, the pose estimator \cite{viola2003fast,wu2007cluster,wu2004fast,zhu2012face} or the 3D object model \cite{hoiem20073d,razavi2010backprojection,seemann2006multi,thomas2006towards,zhang2006generative,savarese20073d}, which make them computationally expensive and require an intensive training process on large datasets. These top-down approaches will result in considerable runtime complexity.

Let's focus on one surveillance camera used in one specific scenario. A common strategy is to train specific object detector from human collected and labelled training samples the scenario since each individual has limited variant poses in one scene. However, it is impossible to train scene-specific detectors for every scenario considering tedious human efforts and time costs, unless the training process requires minimal or even no supervision.

A method must be found to learn a scene-specific object detector without human intervention, extending to other scenes, making sure that every scene has its own detector and achieves satisfactory detection performance under different imaging conditions and viewpoints. Although the idea sounds attractive, this task is challenging because constructing an object model without prior knowledge is difficult, and there is no effective algorithm to automatically collect and label samples for training the detector on the fly. 

Some studies on online-learning object detectors have been proposed and most of them adopt a similar framework, including an online learning detector and a validation strategy. However, these methods are not completely unsupervised learning methods, but only minimize manual effort, and are always initialized by several hundred human labeled training samples for one specific scene. Moreover, the number of manually labeled initial instances will increase proportionally considering the multi-scene object detection application.  In addition, these approaches employ co-training methods \cite{javed2005online,qi2011online}, the background subtraction \cite{nair2004unsupervised}, the generative model \cite{roth2005line,wang2012detection} and tracking-by-detection appoaches \cite{sharma2012unsupervised,sharma2013efficient} as the validation strategies to collect and label the online learning samples automatically. The performance of online-learning approaches are far from competitive with supervised methods because of the high label error in hard samples distributed around the decision hyperplane, compared to the human label in supervised training process.

Recently, some scene-specific object detection approaches \cite{Ye2017Self,Xiao2016Track,Kwak2015Unsupervised} have been proposed to discover and label hard proposals for training an initial generic detector without any manual annotation. Among these various solutions, Ye \cite{Ye2017Self} has shown impressive results under challenging situations by optimizing a progressive latent model (PLM) with the difference of convex (DC) objective functions. However, it is a computational challenge to optimize a DC function straightforwardly throughout the sample space with an unsupervised manner. Furthermore, a careful initial generic detector is necessary to avoid local minima.

To overcome the limitations of existing works, in this paper, a Generative-Discriminative model (GDM) is proposed to partition the sample space into three disjoint subspaces: positive, negative and hard sample space. A Generative model learns the joint probability of positive and negative samples, while hard samples are classified by a discriminative model with solving a mixed integer programming problem. Similar to the concave-convex programming procedure, the GDM can be iteratively updated by an online gradual optimized process which is insensitive to initialization.

Our goal is to design an self-learning object detection framework, which can train object class detector in each particular scenario without human intervention. Instead of manually labeling several hundred initial training samples or employing a generic detector in existing methods, our scene-specific detector is obtained by simply marking several bounding boxes around the object in the first video frame via a mouse. This approach reduces human annotation effort to an effortless mouse operation within the first frame, which "determine" the interested object category in current surveillance video. Other than this, human labeled samples or general object detectors are not needed.

There are two processes in our framework:

In learning process, first, an initial sample set generated by affine transformation of these marked objects in the first frame. Second, a Generative-Discriminative model, trained by the initial sample set, runs as an initial detector on subsequence frames. Obviously, the initial detector has poor detection performance due to the incomplete initial training sample set. Third, for this case, we propose an online gradual optimized procedure to iterative optimize the Generative-Discriminative model (GDM) in an unsupervised manner. When the convergence condition is satisfied, the optimized process will stop and result in a hybrid classifier composed of a generative model and a discriminative model.

During object detection process, the two models work together to determine the location of real objects. The generative model is first used to detect objects based on the sliding window strategy, and the detection responses, located near the classification boundary, will be further recognized by the discriminative model. 

Our method, triggered by several bounding boxes, is minima supervised without human effort on collecting and labelling samples, and can be easily extended to other scenes, forming scene-specific objectors in various surveillance cameras with different viewing distances and angles. Thus, this is a bottom-up method to resolve the multi-scene object detection problem.

Moreover, both the use of Generative-Discriminative model (GDM) and online gradual optmized process make our method robust in tackling the most informative samples lying at the decision boundary, and the method achieves state of the art detection performance under different viewing angles, as shown in Fig. 1. By self-learning within three hours, our method produces satisftying results in three different viewpoints sequences of CAVIAR \cite{sharma2013efficient} and PETS2009 \cite{ferryman2009pets2009} datasets. In addition, we obtain competitive detection performance in the S2 sequences of PETS2009 dataset, compared to the Aggregated Channel Features (ACF) \cite{dollar2014fast}, one of the most robust supervised object detection approaches, trained by manually labeling 300 positive samples and 900 negative samples in the same viewpoint. To the best of our knowledge, this is the first time to demonstrate a scene-specific detector without additional manual annotated samples or generic detectors, compared to other object detection methods.

\begin{figure}[!h]
\centering
\includegraphics[scale=0.4]{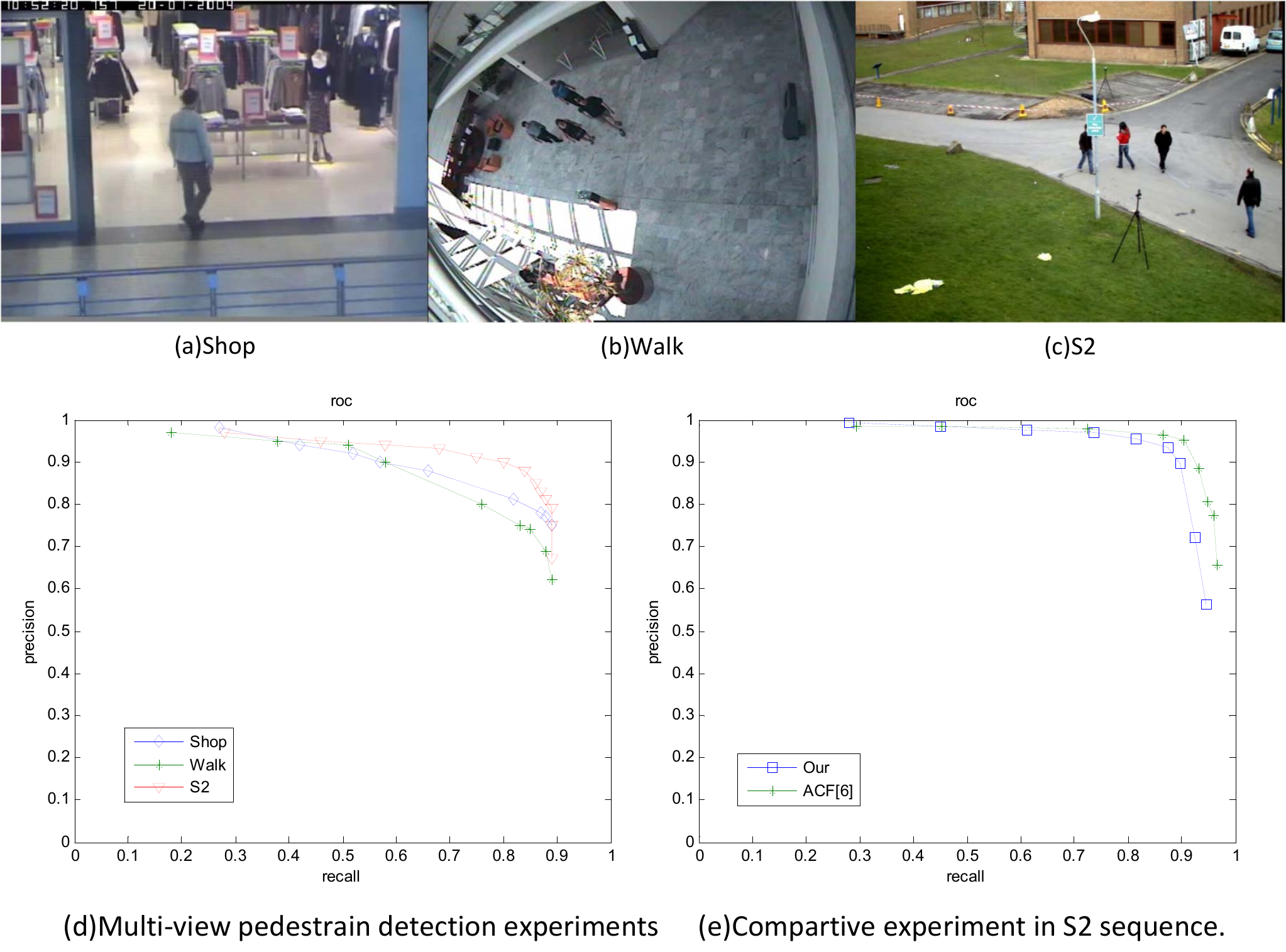}
\caption{(a) Shop sequence from CAVIAR dataset; (b) Walk sequence from CAVIAR dataset; (c) S2 sequence from PETS2009 dataset (d) Experiments in multi-view video sequences; (e) Comparison with ACF[6].}
\label{fig_sim}
\end{figure}

The main contributions of this paper are:

1) We present the Generative-Discriminative model (GDM), which partitions the sample space for improving the discriminability of scene-specific object classifiers while being efficient at run-time.

2) We present the online gradual optimized process which allows the detector, initialized by marking several bounding boxes around the object with poor detection performance, to successively improve classification accuracy and becoming more dedicated to challenging samples lying near the decision boundary.

3) We present a novel self-learning framework to train scene-specific object detectors by marking several bounding boxes around the object in the first frame, which can easily extend to various surveillance videos, and automatically achieve successful detection performance.

The rest of this paper is arranged as follows: Section 2 briefly recalls related works. In Section 3, the analysis of our approach is provided. Generative model is described in section 4, discriminative model is explained in section 5, and section 6 shows the online gradual optimized process. Experiments and results are presented in section 7, which is followed by the conclusions.

\section{Literature Review}

Object class detection is the core component in most computer vision tasks, and it has achieved prominent success for supervised learning based approaches, such as \cite{viola2001rapid,dalal2005histograms,wang2009hog,
felzenszwalb2010object,li2013component,dollar2014fast,Redmon2016YOLO9000}. Here, we focus on the scene-specific object detection, online learning frameworks and multi-view object detection methods.

\subsection{Scene-specific Object Detection}
This paper proposes to address multi-scene object detection problem by bottom-up scene-specific object detection approaches. In essence, \cite{Ye2017Self,shu2013improving,cinbis2017weakly} have been widely used to train the scene-specific detector based on unsupervised object discovery techniques. These self-learning methods usually optimize the convex or semi-convex objective fuctions throughout the sample space. \cite{Ye2017Self} even optimizes the difference of convex objective functions to improve the detection precise with a progressive latent model (PLM).

Despite the success of unsupervised object discovery techniques to improve the performance of the self-learning object detector, they are limited in two aspects. First, this is a mixed integer programming problem which poses a computational challenge when optimizing loss functions directly in the entire feature space with an unsupervised manner. Second, an effective initial generic detector is necessary to avoid falling into the local minima in optimized process. 

\subsection{Online learning Framework}
Online learning frameworks have been published to adapt the detector to a particular scenario. However, it is difficult to use these approaches in the multi-scene object detection domain due to two major reasons:

First, traditional online learning object detection methods are inseparable from manual efforts. For instance, most of the online learning detectors are initialed by human labelled training samples. The detection performances are usually degraded when the number of manual labelled instances are minimized. Some works \cite{wang2012detection,shu2013improving} are proposed to specify a well-trained generic object detector to a specific scene. However, these works are particularly valuable in constrained applicant conditions. Recently, semi-supervised learning \cite{levin2003unsupervised,yang2013semi}, transfer learning \cite{wang2012transferring,pang2011transferring}, and weak-supervised learning \cite{prest2012learning,cinbis2017weakly,liu2016soft,kumar2016track}, are employed to reduce the amount of labeled training data for object detector. How to minimize human effort in online object detection system is still a hot research topic.

Second, the new samples, which are used to train the detector online, need to be automatically collected and labeled. How to correctly label the new training samples is still a challenging topic. To date, various automatic annotation methods have been reported and can be broadly divided into four categories: 1) co-training based, 2) background subtraction based, 3) generative model based, and 4) tracking based.

In co-training based approaches \cite{javed2005online,qi2011online}, two classifiers are trained simultaneously and labeled for each other. In background subtracted approaches \cite{nair2004unsupervised}, the foreground detector, based on a background model, is employed as an automatic labeler. Generative model based methods \cite{roth2005line,wang2012detection} use the reconstructed model error to validate the detection responses and train the detector by a feedback process. Tracking based approaches \cite{sharma2012unsupervised,sharma2013efficient} collect and label online training samples by using tracking-by-detection methods which can interpolate missed object instances and false alarms as positive and negative samples, respectively.

However, the aforementioned methods have no special strategy to deal with the problematic samples located around decision boundary, the most informative and ambiguous part of the object categories. Our method employs online gradual optimized process and the Generative-Discriminative model to hierarchically process the detection responses, which makes the learning procedure focus on the hard samples and reduce the online labeling error.

\subsection{Multi-view Object Methods}
Conventional object detection methods locating objects by considering single view cannot be employed in multi-view object detection since objects have wide variations in their poses, colors and shapes under multi-view imaging conditions. Thus, a common idea is to model object classes by collecting distinct views forming a bank of viewpoint-dependent detectors which can be used to predict the optimal viewing angle for detection.

In early stages, most multi-view detection approaches independently apply several single-view detectors and then combine their responses via arbitrary logic. Some impressive works \cite{fan2005fast,ng1999multi,weber2000viewpoint,li2004floatboost} have been reported in the domain of face detection, dealing with multiple viewpoints (frontal, semi-frontal and profile).

Following this progress, Thomas et.al. \cite{thomas2006towards} no longer rely on single-view detectors working independently, but develop a single integrated multi-view detector that accumulates evidence from different training views. Several other approaches, such as \cite{zhang2006generative,savarese20073d}, build complex 3D part models containing connections between neighboring parts and overlapping viewpoints which achieve remarkable results in predicting a discrete set of object poses. The discrete views are usually treated independently, however, \cite{stark2010back,lopez2011deformable,gu2010discriminative,christoudias2008unsupervised}  require evaluating a large number of view based detectors, resulting in considerable runtime complexity.

Recently, Pepik et.al. \cite{pepik2015multi} proposed 3D deformable part models which extend the deformable part model to include viewpoint information and part-level 3D geometry information. This method establishes represented 3D object parts and synthesizes appearance models for viewpoints of arbitrary granularity on the fly, resulting in significant speed-up. Xu et.al. \cite{xu2015multi} proposed to accomplish multi-view learning with incomplete views by exploiting the connections between multiple views, enabling the incomplete views to be restored with the help of the complete views.

With respect to all the aforementioned approaches, our framework takes a markedly different direction: we establish scene-specific detectors in an unsupervised manner for each scenario, so as to prevent from integrating complex models and labeling substantial training samples in different viewpoints. To resolve the multi-scene object detection problem in bottom-up fashion is an improvement over other systems.

Other than the above mentioned approaches, there exist methods \cite{celik2009unsupervised,huerta2015combining} to detect unknown object classes from the motion segmentation. Although these methods learn foreground models in arbitrary scenarios without any a priori assumption, they are very different from our method, which address one object category detection problem: these methods can not recognize the detected responses because they use the cluster based global optimization procedure.

Our method initializes a scene-specific detector with several bounding boxes in the first frame, and eventually realizes a state of the art multiple object detection system without human label effort. Online gradual optimized strategy, which  is proposed in this paper, is the key point which ensures our framework can be improved from a poor detection performance initial detector (see detail in section 6). Otherwise, the output of our framework is a Generative-Discriminative model which is very different from the other scene-specific object detectors. In this way, our method opens up the possibility for several different classifiers to work together to determine the locations of one object class in videos from self-learning.

\section{Analysis of Our Method}
Object detection can be viewed as a two-class classification problem \cite{amit20022d}. However, in most applications, the sample space can be divided into three groups: positive, hard and negative samples, by two decision boundaries. Both hard positive and hard negative examples have a significant effect on enhancing the classification performance.

In this section, a Generative-Discriminative model has been proposed to describe the three sample spaces from one surveillance domain. A novel cost function will be employed to improve the model performance and speed up the convergence rate. After that, an example of the Generative-Discriminative model will be introduced to learn scene-specific objector with minimal supervision.

\subsection{Generative-Discriminative Model (GDM)}
A generative classifier $ G $, with the inputs $ x $ and the label $ y $, concentrates on capturing the generation process of $ x $ by modelling, which is robust to partial occlusion, viewpoint changes and significant intra-class variation of object appearances. Thus, the model is suitable to describe the positive sample space $ P_{c_{+}} $ and negative sample space $ P_{c_{-}} $.

A discriminative model $ D $, on the other hand, learns the difference between different categories. Thus, $ D $ can be employed to  model the hard sample space $ P_{c_{h}} $, so as to find the optimal classification of samples located between the positive and negative decision boundary: $ B_{+} $, $ B_{-} $.

In other words, samples falling in $ P_{c_{+}} $ and $ P_{c_{-}} $ belong to labeled samples and hard samples are unlabeled samples. Thus, the cost function is as follows:

\begin{equation}
\begin{split}\label{model overall cost function}
L(x)= &\Upsilon\sum_{g(x)>B_{+},g(x)<B_{-}}C_{G}(x,y)+\alpha\sum_{B_{-}\leq g(x)\leq B_{+}}C_{D}(x)\\
\end{split}
\end{equation}

where $ C_{G}(x,y) $ and $ C_{D}(x) $ are the objective functions for labelled samples and unlabelled samples respectively. $ \Upsilon $ and $ \alpha $ are regularization factors. $ g(.) $ is the classification function calculated by Generative model.

In addition, the distance $ Dis(B_{+},B_{-}) $ between the positive and negative decision boundaries has a huge impact on the model. The smaller the distance, the more accurate the generative model is to describe the positive and negative samples. Furthermore, a smaller distance also means fewer hard samples, which provides a convergence condition in optimizing a Generative-Discriminative model. Thus, the cost function can be formulated as follows:

\begin{equation}
\begin{split}\label{model cost function}
L(x)= &\Upsilon\sum_{x\in P_{c_{+}},P_{c_{-}}}C_{G}(x,y)+\alpha\sum_{x\in P_{c_{h}}}C_{D}(x) +\lambda \cdot Dis(B_{+},B_{-})
\end{split}
\end{equation}

where $ \lambda $ is the regularization fact of the distance item $ Dis(.) $. 

The cost function $ L(x) $ has three latent variables: the pseudo labels of unlabelled samples, the positive boundary $ B_{+} $ and the negative boundary $ B_{-} $, which cannot be solved efficiently with the traditional tools of optimization. In this paper, we propose the online gradual optimized process to solve the objective function in a dynamically changing sample space (see detail in section 6).

Many generative and discriminative models can be employed in the GDM, such as Autoencoder networks, conditional random fields, hidden Markov model and Bayesian networks. In our case, we propose an Online Selector Fern(OSF) generative model and an iterative SVM(ISVM) discriminative model, considering real-time requirements in video object detection applications. This results in a hybrid model which is both highly effective and computationally efficient (running at over 60 fps for $ 768\times576 $ images).

\subsection{Self-learning Framework}

Based on the Generative-Discriminative model, a self-learning framework is proposed to train a scene-specific object detector. The initial training sample set is prepared by affine transformation of the several selected patches in the first surveillance video frame. Thus, the human labeled training data or a generic detector are not needed. An OSF generative model is trained by the initial training data as the initial detector. Obviously, the detector has poor detection performance due to the incomplete training sample set. As shown in Fig. 2, the detector will be improved by a propagation optimal process.

Thus, we propose the online gradual optimized process to train a Generative-Discriminative model in fully-autonomous fashion:

\begin{figure*}[htbp]
\centering
\includegraphics[scale=0.4]{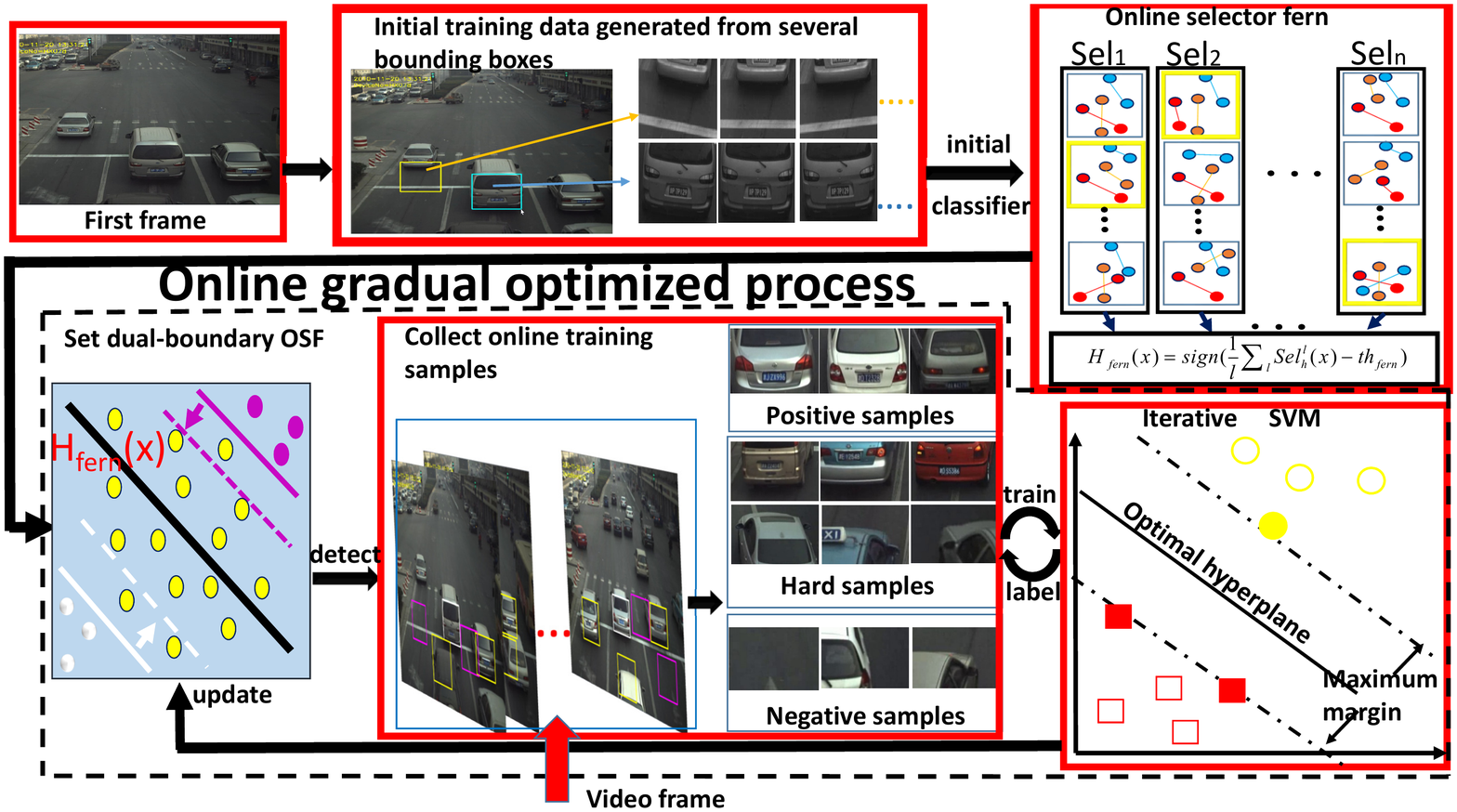}
\caption{Approach overview}
\label{fig_sim}
\end{figure*}

First, we define positive and negative decision boundaries in the OSF classifier. The detected responses are then collected as positive, negative and hard samples, respectively. To ensure the high correct label rate, the OSF classifier has large initial margin setting between the two boundaries.

Second, an ISVM discriminative model is proposed to solve a mixed integer programming problem and label the hard samples in the dynamically changing subspace.

Third, the labelled hard samples are used to online train the OSF classifier and gradually minimize the margin between positive and negative boundaries, improving the ability to express the positive and negative sample spaces.

This process is repeated till convergence. The output is a Generative-Discriminative model, composed of an OSF classifier and an ISVM model.

In detection process, the scanning window strategy is employed. Most image patches are classified by the OSF classifier and only a small fraction located between the dual boundaries, collected as hard samples, are classified by the ISVM model, which makes our approach robust while being efficient at run-time.

Next section, we will first introduce the Boosting fern classifier and its online learning algorithm realized by an online selection operator. The iterative SVM will be presented in section 5. Following, the online gradual optimal process is shown in section 6.

\section{Online Selector Fern(OSF) Generative Model}
Traditional fern classifiers are widely used in the object tracking field \cite{ozuysal2007fast}, due to efficiency and high performance in tracking affine transformation planar objects. \cite{villamizar2010efficient,villamizar2012bootstrapping,levi2013fast} extended fern classifier methods to detect objects appearing in the image under different orientations and view points. In this paper, we propose the online selector fern (OSF) integrated fern classifiers and an online feature selection strategy. Fern classifiers play the role of weak classifier and can be boosted into a strong model by online selection operators. Furthermore, the generative model will evolve into the dual-boundaries OSF, when the positive and negative boundaries are initialized around the decision hyperplane of the generative model.

First, we briefly describe the boosting fern model \cite{villamizar2012bootstrapping}, second, the OSF generative model will be introduced to further improve the ability of the model to express labelled samples, and third, the positive and negative boundaries will be initialized around the decision hyperplane of the OSF model, which evolves into the dual-boundary OSF classifier with the least model parameters.

\subsection{Boosting Fern}
Let $ S^{i}=(f^{i},C^{i}), i=1,2,...,I $ denotes training samples, which is an image patch set and their labels. Where $ I $ is the number of samples, $ C\subseteq \lbrace c_{+},c_{-}\rbrace$ is the sample labels, $ f $ is an J-dimension feature vector which is used to describe the samples:
\begin{equation}\label{Fernfeature}
f=(f_{1},f_{2},...,f_{J})
\end{equation}

We employ the Local Binary Feature (LBF) \cite{ozuysal2007fast} to map an image sample to a Boolean feature space by comparison between two random position intensities of a sample. Fern is denoted as a feature sub-space random sampled from the $ J $-dimension feature space. The $ l $th fern can be denoted as:

\begin{equation}\label{FernCla}
F_{l}={f_{l,1},f_{l,2},...,f_{l,s}},\quad l=1,2,...,L
\end{equation}

Where $ s $ is the number of features in a fern. For a training sample, this gives an $ s $-digit binary code to describe the sample appearances. In other words, each Fern maps 2D image samples to a 2$ s $-dimensional feature space, as shown in Fig. 3 (a). Accordingly, apply the fern $ F_{l} $ to each labelled training sample and learning the posterior distribution as histogram in each class $ P(F_{l}|c_{+}) $ and $ P(F_{l}|c_{-}) $, as shown in Fig. 3(b).
\begin{figure}[!h]
\centering
\includegraphics[scale=0.30]{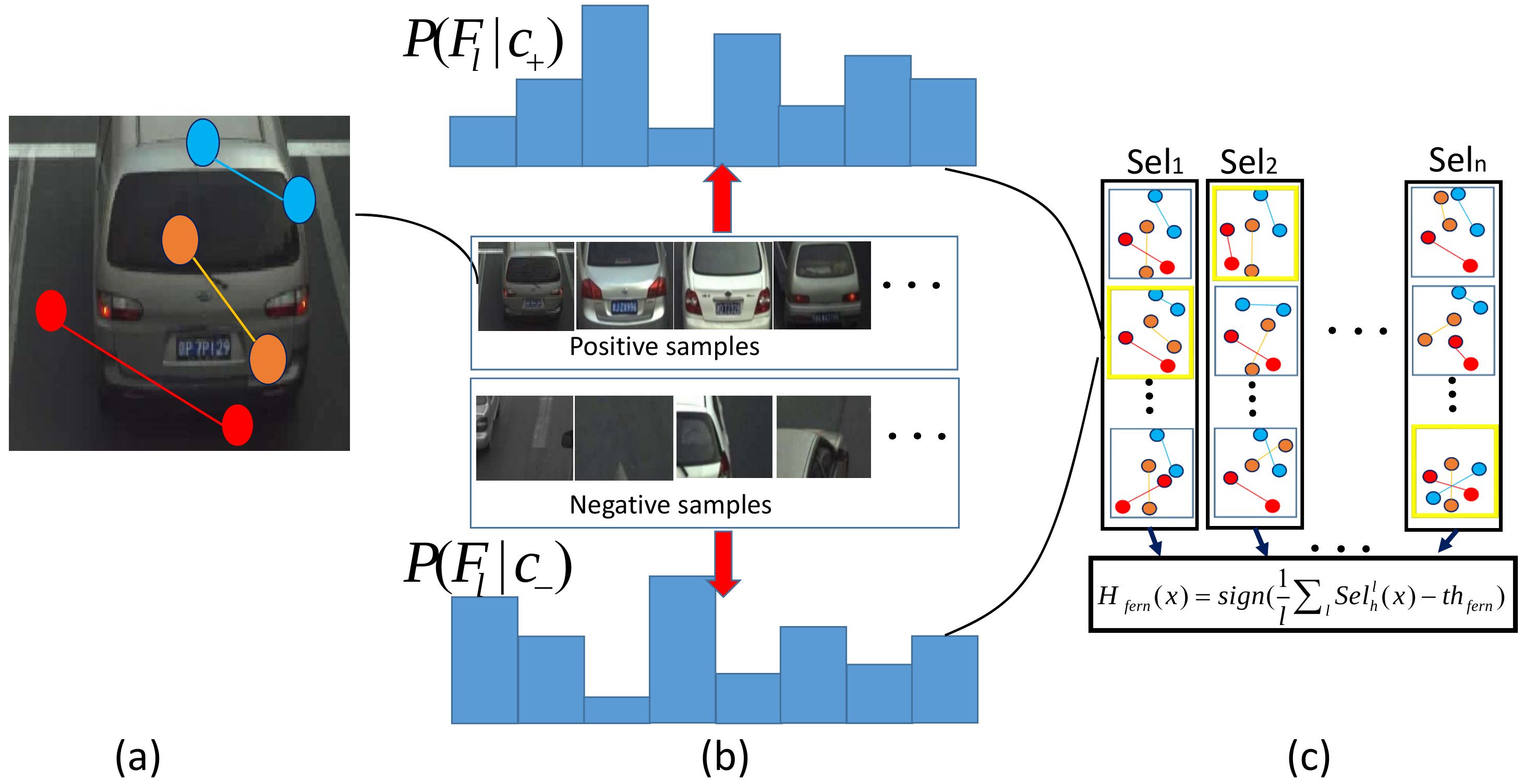}
\caption{(a) LBF features;(b) posterior distribution of $F_{l}$;(c) online selector of Boosting fern.}
\label{fig_sim}
\end{figure}

A single fern does not give the accurate estimation of the generative model in a specific surveillance scene, but we can build an ensemble of ferns by randomly choosing different subsets of LBF features. A weak classifier is defined by the co-occurrence of a random Fern $ F_{l} $'s observation, which is further linearly combined to an ensemble classifier.

\begin{equation}\label{FernOber}
h_{l}(x)=\dfrac{P(F_{l}(x)=k|c_{+})+\varepsilon}{P(F_{l}(x)=k|c_{+})+P(F_{l}(x)=k|c_{-})+\varepsilon}
\end{equation}

where $ \varepsilon $ is a smoothing factor. Thus, the classifier is estimated approximately:

\begin{equation}\label{Ferneval}
H_{fern}(x)=sign(\frac{1}{L}\sum_{l}h_{l}(x)-th_{fern})
\end{equation}

$ th_{fern} $ is the threshold of each weak fern classifier, which is usually set to 0.5. Next section will introduce the weak classifier selection strategy and the online selector fern algorithm.

\subsection{Online Selector Fern(OSF)}	

From section 4.1, the fern based weak classifier is more discriminant than single feature based weak classifier because every fern is a group of features in fixed size of $ s $. Moreover, the online learning process of each weak classifier is simplified by updating the post probability of every fern. However, in the online learning process, we must find a method to select the most discriminant fern classifier and minimize the first item in equation (2), the cost function of a Generative-Discriminative model. We denote a fern based weak classifier set $ h_{l}^{set}={h_{F_{1}}(x),h_{F_{2}}(x),h_{F_{3}}(x),...,h_{F_{M}}(x)} $, and a selector:
\begin{equation}\label{Fernsel}
Sel_{l}= h_{F_{m}}(x),m\in {1,2,...,M}
\end{equation}

where $ m $ is chosen according to an optimization criterion. In this paper, we use the criterion of picking the one that minimizes the following Bhattacharyya distance between the distributions of the class $ c_{+} $ and background $ c_{-} $:

\begin{equation}\label{Ferncriterion}
B_{l}=2\sum_{k=0}^{2^{r}}\sqrt{P(F_{l}(x=k)|c_{+})P(F_{l}(x=k)|c_{-})}
\end{equation}

As shown in Fig. 3(c), a fixed set of $ N $ selectors are initialized randomly, each has a fixed set of $ M $ fern based weak classifiers. When the weak classifiers of each selector receive a new training sample, the classifiers are updated by changing the post probability distribution of each fern according to the location in a $ 2^{s} $ dimension feature space partitioned by the fern. The weak classifier with the smallest Bhattacharyya distance is selected. This procedure is repeated for all selectors. A strong classifier is obtained by linear combination of selectors.

\begin{equation}\label{StroFern}
H_{fern}(x)=sign(\frac{1}{L}\sum_{l}Sel_{h}^{l}(x)-th_{fern})
\end{equation}

\subsection{Dual-boundaries OSF Classifier}

Let $ H_{fern}^{0}(x) $ denote the initial OSF classifier based detector, which is firstly applied to the target video by sliding window search. All detected visual examples $ d_{i}(i\in {1,2,...,N}) $ are collected and divided into positive sample set, $ Set_{pos} $, hard sample set, $ Set_{hard} $ and negative sample set, $ Set_{neg} $ based on the confidence level calculated by the OSF classifier.

\begin{equation}\label{EquGradual}
\left\{
\begin{split}
&d_{i}\in Set_{pos}\quad if\quad H_{fern}^{0}(d_{i})>\beta +\dfrac{\theta}{2}\\
&d_{i}\in Set_{hard}\quad if\quad \beta +\dfrac{\theta}{2}>H_{fern}^{0}(d_{i})>\beta -\dfrac{\theta}{2}\\
&d_{i}\in Set_{neg}\quad if\quad H_{fern}^{0}(d_{i})<\beta -\dfrac{\theta}{2}
\end{split}
\right.
\end{equation}

$ \beta $ is the decision hyperplane. $ \beta+\dfrac{\theta}{2} $ and $ \beta-\dfrac{\theta}{2} $ are the positive and negative boundaries around $ \beta $ and have large margin in the initial stage. Thus, the most detection responses will located between the two boundaries and collected as hard samples with uncertain labels. The online selector fern classifier becomes an dual-boundary OSF.

Simplifying the parameters about the positive and negative boundaries to a scalar threshold $ \beta $ greatly reduces the complexity of the GDM. Experiments show that the model performs well even in clutter surveillance videos, as shown in Fig. 5. 

\section{Iterative SVM}
A standard SVM classifier for two-class problem can be defined as:
\begin{equation}\label{SVM}
\left\{
\begin{split}
&\mathop{min}\limits_{W,b,\xi} \ \dfrac{1}{2}||W||^{2}+c\sum_{i=1}^{N}\xi_{i},\\s.t.\forall i:\
&y_{i}(W^{T}x_{i}+b)\geq 1-\xi_{i}, \xi_{i}\geq 0,i=1,2,...,N
\end{split}
\right.
\end{equation}

where $ x_{i}\in R^{n} $ is a training sample, $ y_{i}\in \lbrace -1,1\rbrace$ is the label of $ x_{i} $. $ c>0 $ is a regularization constant.

The traditional SVM belongs to the supervised discriminative model. However, semi-supervised SVM \cite{shah2008svm} or mi-SVM (Multiple Instance SVM ) \cite{Andrews2003Support} can be trained by labelled and unlabelled samples simultaneously. In this paper, labelled samples, collected from several bounding boxes in the first frame, are much less than the traditional semi-supervised algorithms. Therefore, an iterative SVM algorithm is proposed to gradually label high-confidence hard samples in subspace. 

As shown in Fig. 2, the initial training samples, generated by affine transformation of the several bounding boxes in the first frame, can be used as the initial labelled sample set $ L_{0}={(x_{1},y_{1}),(x_{2},y_{2}),...,(x_{m},y_{m})} $, and the hard samples takes the place of unlabeled sample set $ U={x_{n},x_{n+1},...,x_{g}} $ which is automatically collected from the detection responses of online selector fern based detector. 

\begin{equation}\label{SVM}
\left\{
\begin{split}
&\mathop{min} \limits_{\hat y_{j}}\mathop{min}\limits_{W,b,\xi} \ \dfrac{1}{2}||W||^{2}+c\sum_{i=1}^{N}\xi_{i}+d\sum_{j=1}^{M}\xi_{j}, \\s.t.\forall i:\
&y_{i}(W^{T}x_{i}+b)\geq 1-\xi_{i}, \xi_{i}\geq 0,x_{i}\subseteq L, \\s.t.\forall j:\
&\hat y_{j}(W^{T}x_{j}+b)\geq 1-\xi_{j}, \xi_{j}\geq 0,x_{j}\subseteq U
\end{split}
\right.
\end{equation}

where $ \hat y_{j} $ is the pseudo labels of unlabelled samples, and $ d $ is the regularization constant of the unlabelled term.

Thus, the manual labelled samples are no longer needed in the optimized process. The SVM model can be trained in an unsupervised manner, as follows:

First, the initial SVM classifier, denoted as $ H^{0}_{SVM}(x) $, is trained by the same initial sample set, generated by affine transformation of the bounding boxes in the first frame, which ensures the model is correctly initialized. The HOG feature \cite{dalal2005histograms} is employed to train $ H^{0}_{SVM}(x) $, which is different from the LBF feature used in the fern classifier training process. Thus, two different types of features can be integrated in our self-learning framework, which is crucial to improve the detection performance in cluttered environments \cite{wang2009hog}.

Second, we perform the $ H^{0}_{SVM}(x) $ on the unlabeled sample set $ U $. The predicted labels are denoted as $ lab_{u}={y_{n}(0),y_{n+1}(0)},...,y_{g}(0) $. 

Third, denote $ th^{P}_{SVM} $ and $ th^{N}_{SVM} $ as the positive and negative thresholds according to SVM classification confidence. The labelled positive sample set can be updated by adding some hard samples with high classification confidence (more than $ th^{P}_{SVM} $), while the labelled negative sample set can be updated by adding some hard samples with low classification score (less than $ th^{N}_{SVM} $). This is a  more conservative manner to update the labelled sample set when compared to the convenient semi-supervised SVM algorithm. Using the updated training set $ L_{t} $, we train a new  model, and perform classification again on $ U $. The predicted labels are denoted as $ lab_{u}={y_{n}(1),y_{n+1}(1)},...,y_{g}(1) $.

If all the hard example labels are unchanged, the algorithm stops after the $ k $th iteration. The SVM model and the predicted labels $ lab_{u}={y_{n}(k),y_{n+1}(k)},...,y_{g}(k) $ of the hard sample set are the final output. If changed, perform the second and third step for the $ (k+1) $th iteration.

This is a heuristic optimization procedure similar to mi-SVM \cite{Andrews2003Support} without the constraint of at least one positive sample in the unlabelled sample set. In addition, the optimization process performs in a subspace, embedding in the online gradual optimal process, produces surprising experimental results without fine tuning, as shown in Fig. 8.

The iterative optimized process is triggered by an online hard sample collected module without manual annotation, as shown in Fig. 2. When the online selector Ferns are retrained by the convergent iterative SVM and labeled hard sample set according to a online gradual optimized process, we obtain a Generative-Discriminative model, consisting of a $ H_{fern}(x) $ and a $ H_{SVM}(x) $, more dedicated to the problematic samples. As a consequence, the updated classifier puts more emphasis on the most distinctive parts of the object that reduces the global classification error.

\section{Online Gradual Optimized Procedure}
With the online gradual optimized procedure, a poor performance detector is acceptable in the beginning of the learning process, and will improve from iteratively learning the hard samples located close to the decision boundary, which is the key idea behind our proposed framework.

The optimized procedure is divided into three steps. The first step optimize the generative model by solving $ C_{G}(x,y) $, the second step labels hard samples by optimizing the discriminative model $ D $, and the third step reduce the margin gradually between the positive and negative boundaries. 

In essence, this procedure similar to an active learning process [56, 57] which reduce gradually the amount of human assistance of annotating hard samples by updating classifier models for the most distinctive parts of the object domain. However, to obtain labels for these samples, the active learner has to ask an oracle (e.g., a human expert) for labels. In our framework, the human expert can be instead by the discriminative model which has learned the hard samples in advance. Thus, the distance $ Dis(B_{+},B_{-}) $ can be updated by evaluating the performance of the generative classifier $ G $ in real time.

From equation (10), the parameter $ \theta $, determining the margin between the dual boundaries, can be minimized by the followed equation:

\begin{equation}\label{DualFernBoundary}
\theta =1-\nu \zeta_{fern}
\end{equation}

where $ \nu $ is a sensitivity parameter which controls the learning speed of dual-boundary OSF classifier (set to 0.85  in our experiments). $ \zeta_{fern} $ measures the performance of the OSF classifier which makes the margin reduce process adaptive to the classifier learning process.  $ \zeta_{fern} $ can be computed by:

\begin{equation}\label{DualFernMeasure}
\zeta_{fern}=\dfrac{\sum_{d_{i}\in Set_{hard}}[(H_{fern}(d_{i})-\beta)\times sign(H_{SVM}(d_{i}))]}{\sum_{d_{i}\in Set_{hard}}|(H_{fern}(d_{i})-\beta)|}
\end{equation}

\begin{table}[htbp]  %%%%
\centering
\caption{ONLINE GRADUAL OPTIMIZED PROCESS}
  \label{tab:ModelASymbol}
 \begin{tabular}{p{0.9\columnwidth}}
 \toprule
\textbf{Input}: The initial training sample set $ L_{0}={(x_{1},y_{1}),(x_{2},y_{2}),...,(x_{m},y_{m})} $ generated from the affine warping of several bounding boxes in the first video frame. Denote empty hard sample set   $Set_{hard}\in\{\}$.Initialled online dual-boundary OSF classifier $H_{fern}^{0}(x)$, and some initialled parameters including $\beta=0.5$, $\theta_{0}=1$ and $\nu=0.85$.  \\
 \midrule
\textbf{Output}: $H_{hyb}(x)$composed by a $H_{fern}(x)$ and a $H_{SVM}(x)$. \\
\midrule
Training initial SVM model $H_{SVM}^{0}(x)$ from the initial training sample set $L0$.\\
t=0 \\
while ($\theta>0.3$) \\
  \quad --using $H_{fern}^{t}(x)$ to detect object in video \\
  \quad --collect $d_{i}\in Set_{hard}$, and calculated the number of collected \\
  \qquad hard samples $N_{hard}$ \\
  \quad --if ($N_{hard} > 100$) \\
    \qquad \\
  \qquad $H_{SVM}^{t+1}(x)$ = ISVM($H_{SVM}^{t}(x)$, $Set_{hard}$, $L_{0}$)\\
  \qquad ; optimize the discriminative model in eq (2)  \\
  \qquad \\
  \qquad $Lab_{u}=H_{SVM}^{t+1}(Set_{hard})$\\
  \qquad ; label the hard samples set by $H_{SVM}^{t}(x)$ \\
    \qquad \\
  \qquad $H_{fern}^{t+1}(x)$= OSF($H_{fern}^{t}(x)$,$Set_{hard}$,$Lab_{u}$)\\
  \qquad ; optimize the generative model in eq (2) \\
    \qquad \\
  \qquad classify hard sample by $H_{fern}^{t+1}(x)$ \\
    \qquad \\
  \qquad calculate $\zeta_{fern}$ by equation (14) \\
    \qquad \\
  \qquad calculate $\theta$ by equation (13) \\
    \qquad ; optimize the distance item in eq (2) \\
    \qquad \\
  \qquad $Set_{hard}\in\{\}$; clear the $Set_{hard}$ \\
    \qquad \\
  \qquad t = t+1  \\
    \qquad \\
  \quad --end if  \\
repeat  \\
 \bottomrule
 \end{tabular}
\end{table}

The overall online gradual optimized process is shown in Tab. 1. The output is a Generative-Discriminative model integrated by a ISVM model and a OSF classifier with dual decision boundaries. When the Generative-Discriminative model is used to detect individual class objects in a video, most candidate windows, generated by the sliding window strategy, are classified by the OSF classifier. Only a small fraction, located between the positive and negative boundaries, are dedicatedly classified by the ISVM model. This hybrid classifier system exploits the strengths of the individual classifier models by first performing a sample space partitioning and, second, assigning to each partition region an individual classifier that achieves high classification accuracy while being efficient at run-time. Moreover, the self-learning process is fully-autonomous and can be extended to other surveillance scenes or object class detection tasks. Resolving the multi-scene object detection problem is very important, and can be settled by combining multiple self-learning scene-specific detectors.

\section{Experiments And Comparisons}

The proposed method is evaluated on vehicle and pedestrian detection problems, which play a key role in current intelligent surveillance systems. For the vehicle detection task, the GRAM-RTM dataset \cite{guerrero2013vehicle} and the Vehicle dataset are used to evaluate our approach.  The Vehicle dataset has been captured and labeled by ourselves. The two datasets, composed of 6 video sequences with different view points and resolution levels, show a real urban road scene with multiple vehicles at the same time.  As shown in Fig. 4, four sequences: Hx, Yk, Hi and Hn, are used from these datasets, which have 6415, 1663, 7520 and 1085 frames, respectively, of different resolutions: $ 1224\times1024 $, $ 512\times288 $, $ 800\times480 $ and $ 2448\times2048 $. Hx has 912 GT instances of the vehicle, whereas Yk and Hi have 344 and 2089 GT instances of the vehicle, respectively. To evaluate the pedestrian detection performance, six sequences: TownCentreXVID.avi (Town), PNNLParkingLot2.avi (PNN), WalkByShop1front.avi (Shop), OneShopLeave2Enter (Enter), Meet-Crowd.avi (Walk), and S2.L1View-001.avi (S2), are used from the well-known public Towncenter \cite{Benfold2011Stable}, PNN-Parking-Lot2/Pizza \cite{shu2013improving}, CAVIAR \cite{sharma2013efficient} and PETS2009 \cite{ferryman2009pets2009} datasets, which have different appearances due to different imaging view points, as shown in Fig. 4. The Ground-Truth is available at \cite{sharma2013efficient,ferryman2009pets2009,Benfold2011Stable,shu2013improving}. Note that the Town sequences are much longer than the other sequences with an average of about sixteen people visible at any frame. This is a challenged dataset to most pedestrian detection methods.
\begin{figure}[!h]
\centering
\includegraphics[scale=0.32]{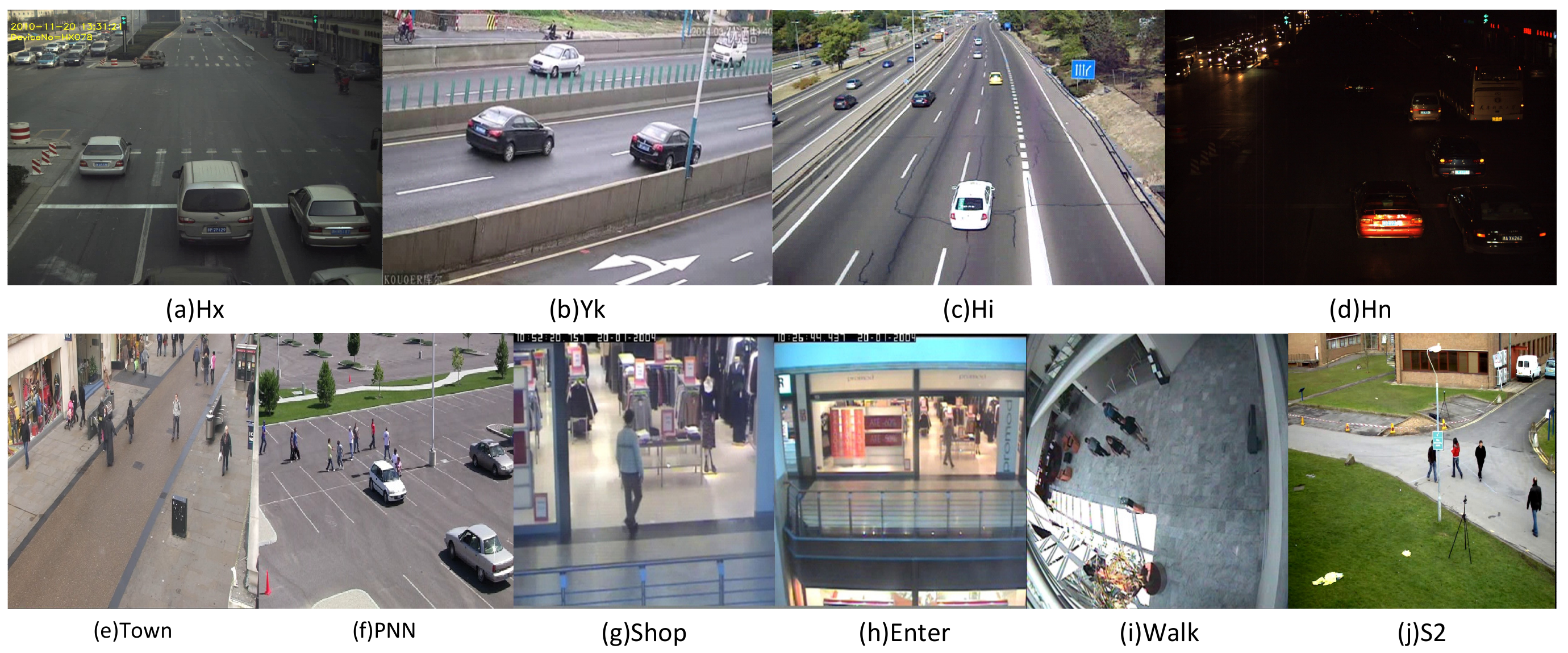}
\caption{Video sequences.}
\label{fig_sim}
\end{figure}

In each experiment, we trigger the video-specific object learning algorithm by several bounding boxes in the first frame. Thus, this method can be conveniently extended into each video sequence. In terms of the necessary parameters that define our Generative-Discriminative model, we would like to point out in all the experiments, described in the following section, the dual-boundary OSF classifier have 10 selectors with initial margin parameter $ \theta_{0} $ setting to 1. Each selector uses 10 Random fern classifiers with 6 binary local features. When online optimizing the ISVM model, HOG feature is employed to describe the samples which are divided into cells of size $ 16\times16 $ pixels and each group of $ 8\times8 $ cells is integrated into a block. In other parameters, in online gradually optimized process, $ \nu $, controlling the updating speed of the online dual-boundary OSF classifier, is set to 0.85.

In traditional object detection methods, the detection performance varies considerably due to resolution difference in different videos. It is difficult for these traditional methods to set a appropriate detection scale to satisfy all the possible resolutions in different videos. But in our framework, we can conveniently ensure the optimal detection scales for every testing video according to the bounding box in the first frame, which describes the accurate object size in surveillance videos. From this, we increase 11 different scales and achieve robust detection performance in each testing video.

In experiments, our approach has demonstrated state of the art detection performance in each testing video after no more than 5 hours self-training process, learning about 400-1500 samples without any human effort, as shown in Tab. 6.

\subsection{Online Generative-Discriminative model Structure and Parameter Test}
\begin{figure}[!h]
\centering
\includegraphics[scale=0.60]{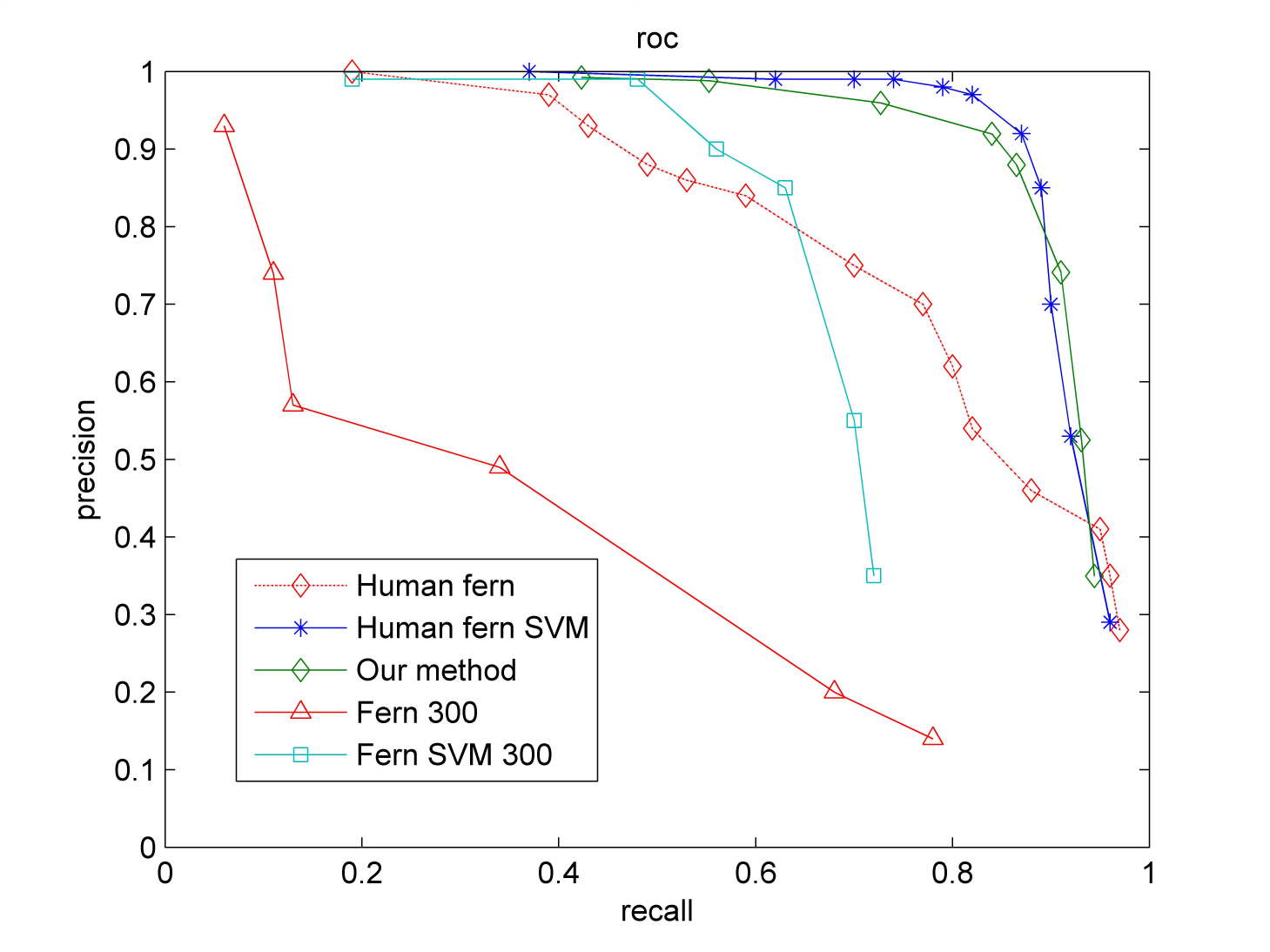}
\caption{Experiments on analyzing our framework structure.}
\label{fig_sim}
\end{figure}

We initially analyze the object detection performance of the proposed framework on the sequence Hx from the Vehicle dataset, which will reveal the influence of different hybrid classifier strategies and human annotation. As shown in Fig. 5, these classifiers were initialized by the same sample set generated by affine warping of several bounding boxes in the first frame of the sequence Hx, but have different learning processes and classifier systems. Here, we evaluate five different alternatives:

\textbf{Human fern}: the vehicles are detected only by the OSF classifier, which is online trained by 850 human labeled  frames, including 246 positive samples and 500 negative samples collected from the sequence Hx.

\textbf{Human fern SVM}: the detector is a Generative-Discriminative model consisting of a dual-boundary OSF classifier and an ISVM model, which is supervised by 850 frames, including manually annotated 246 positive samples and 500 negative samples.

\textbf{Our approach}: The detector is a Generative-Discriminative model composed of a dual-boundary OSF classifier and an ISVM classifier. The OSF classifier is used first to detect objects by the sliding window searching strategy and then the ISVM focuses on recognizing the hard samples distributed around the decision boundary. Note the Generative-Discriminative model is obtained by self-learning \textbf{1460 }frames, about\textbf{ 336 }positive samples and \textbf{687 }negative samples, all collected and labeled automatically; some collected samples are shown in Fig. 6.
\begin{figure}[!h]
\centering
\includegraphics[scale=0.5]{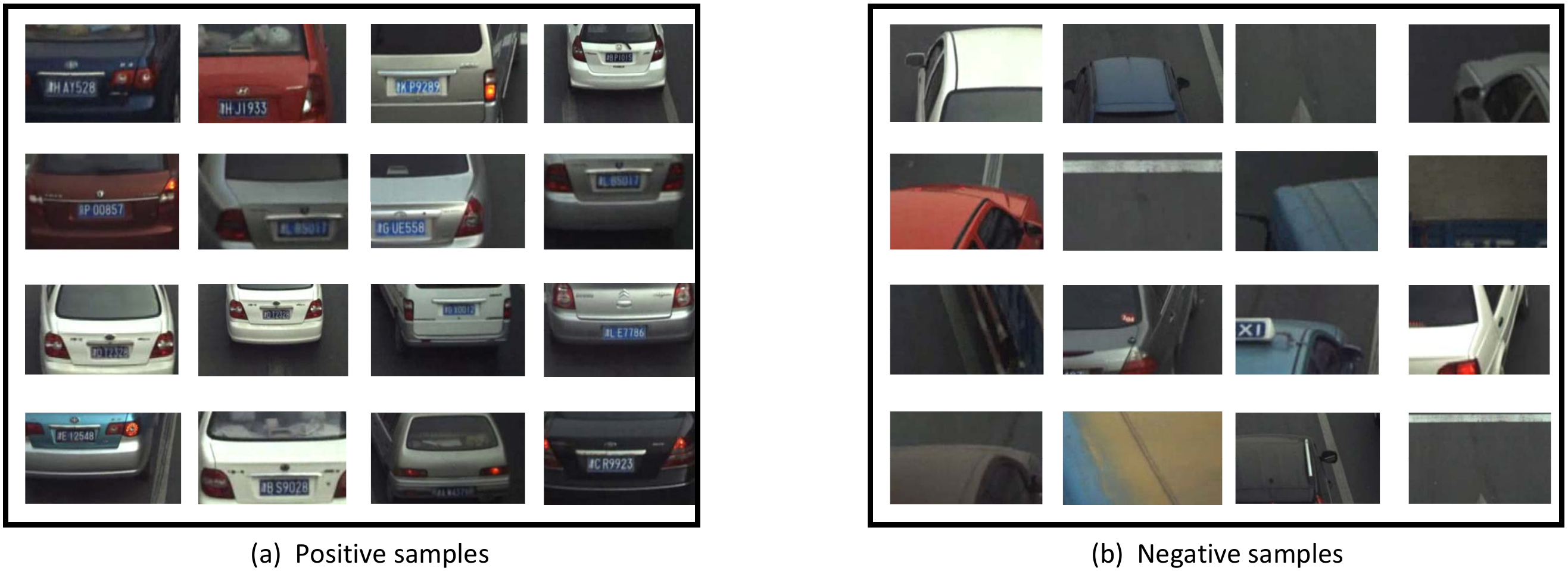}
\caption{Online collected training data.}
\label{fig_sim}
\end{figure}

\textbf{Fern 300}: the detector is the only OSF classifier which is supervised by 300 frames under human guidance, learning about 160 positive samples and 160 negative samples.

\textbf{Fern SVM 300}: the detector is a Generative-Discriminative model which is obtained by human annotated 300 frames, about 160 positive samples and 160 negative samples.

The ROC curves are shown in Fig. 5. The Generative-Discriminative model based detector significantly outperforms single classifier. The results clearly demonstrate the ability of our Generative-Discriminative model to handle the most distinctive parts of the object category.

Our method has comparable performance to the supervised Generative-Discriminative model (Human fern SVM), demonstrating our framework, with a high label correct rate, can improve detection performance by focusing learning on the problematic samples located near the decision boundary.

\begin{figure}[!h]
\centering
\includegraphics[scale=0.6]{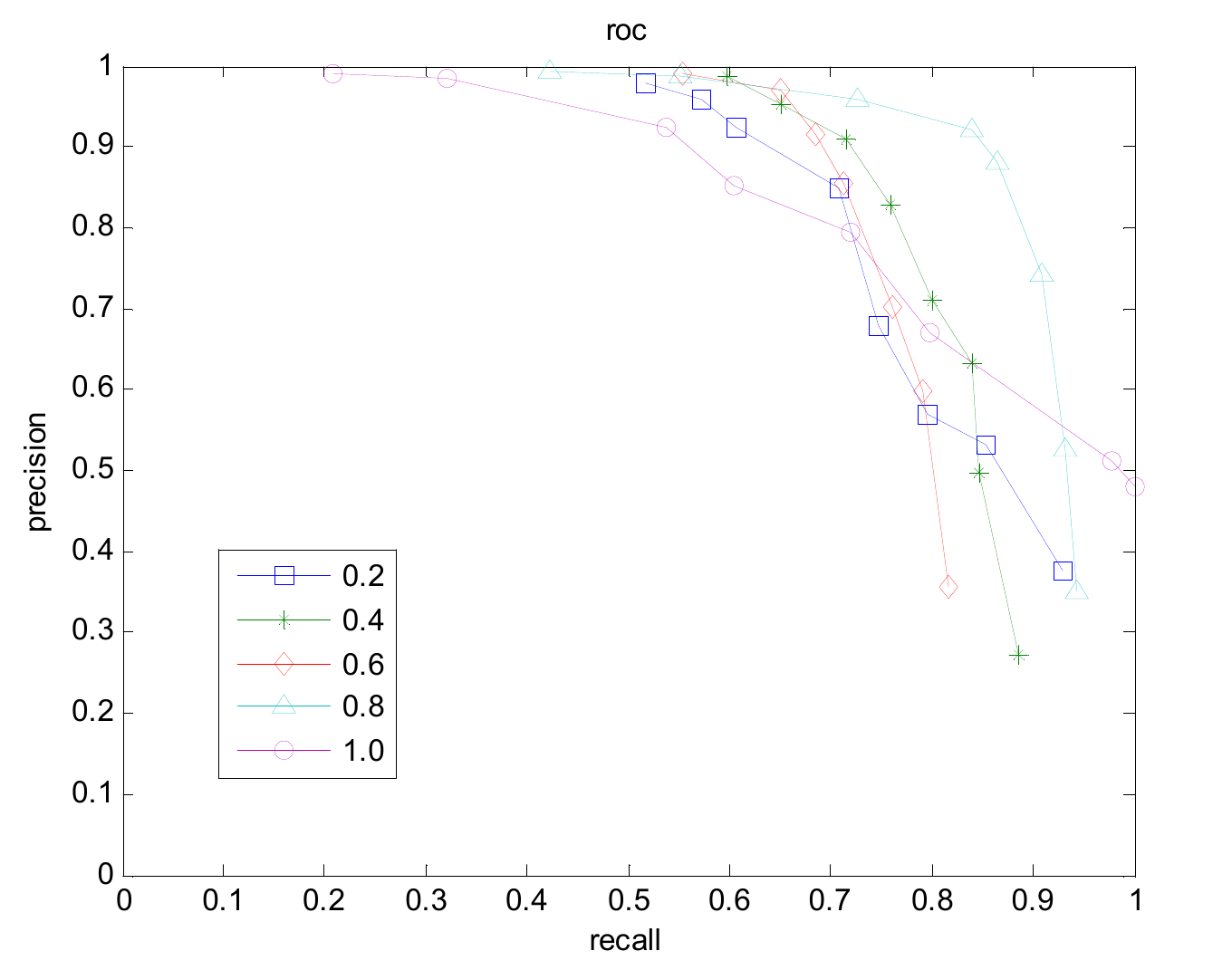}
\caption{Parameter $ th_{p} $ test.}
\label{fig_thn}
\end{figure}

In addition, the positive and negative thresholds $ th^{P}_{SVM} $ and $ th^{N}_{SVM} $, denoted in the ISVM model, are critical for the "hard sample" learning. The two parameters are mutually opposite numbers. We change the value of $ th^{P}_{SVM} $ from 0.2 to 1.0. and applied to vehicle detection on sequence Hx. The ROC curve is shown in Fig. \ref{fig_thn}. The F-measure is calculated in Tab. \ref{thn}. From the experimental point of view, when the $ th^{P}_{SVM} $ is about 0.8 (remains the same in subsequent experiments), the detection performance is satisfactory.

\begin{table}[htbp]
\centering
\caption{COMPARISON WITH DIFFERENT $ th^{P}_{SVM} $}
  \label{tab:ModelASymbol}
 \begin{tabular}{cccccc}
 \toprule
 Sequence & 0.2 & 0.4 & 0.6 & 0.8 & 1.0 \\
 \midrule
 Hx  &0.7718& 0.8010&  0.7795& 0.8779& 0.7544\\
 \bottomrule
 \end{tabular}
 \label{thn}
\end{table}

\subsection{Comparison with Scene-specific object detection Methods}

\begin{figure}[!h]
\centering
\includegraphics[scale=0.5]{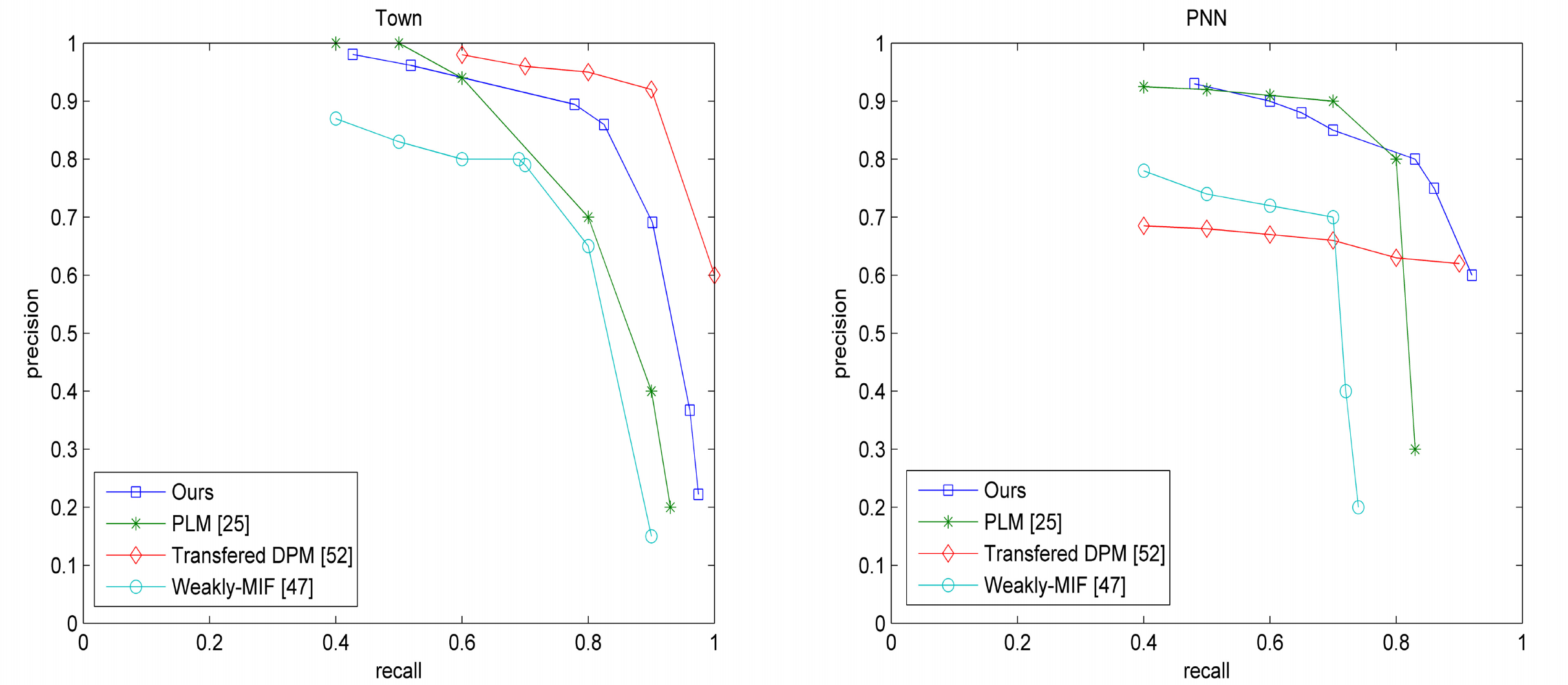}
\caption{Comparison with scene-specific object detection methods}
\label{SceneRoc}
\end{figure}

\cite{Ye2017Self,shu2013improving,cinbis2017weakly} are the widely used scene-specific object detection methods in recent years. \cite{Ye2017Self} achieves the best performance in multiple databases due to the progressive latent model and the graph-based label propagation. Compared with \cite{Ye2017Self}, the GDM improves the precision by nearly 10 percent in the Towncentre dataset, and achieves the best detection performance in the PNN-Parking-Lot2/Pizza dataset, as shown in Fig. \ref{SceneRoc} and Tab. \ref{SceneTab}. The main reason is that our method proposes to optimize a multiple latent variables model in subspace, and can autonomously learn a scene-specific detector through a poor initial detector. Detection results are shown in Fig. \ref{fig_11}.

\begin{table}[htbp]
\centering
\caption{COMPARISON WITH SCENE-SPECIFIC OBJECT DETECTION METHODS}
  \label{SceneTab}
 \begin{tabular}{ccccc}
 \toprule
 Sequence & Shu[47] & Cinbis[52] & Ye[25] & Ours \\
 \midrule
 Town  & 0.7422 & 0.9098 & 0.7466 & 0.8419\\
 PNN & 0.7000 & 0.7342 & 0.8000 & 0.8147\\
 \bottomrule
 \end{tabular}
  \label{SceneTab}
\end{table}

\subsection{Comparison with Online and offline Learning Methods}

In this section, the proposed method is compared with three online learning object detection methods \cite{sharma2013efficient,qi2011online,roth2005line}, which have achieved satisfactory results among the online learning object detection frameworks. More specifically, the proposed approach is compared with four supervised methods: Boosted fern \cite{villamizar2012bootstrapping}, ISVM \cite{shah2008svm}, ACF \cite{dollar2014fast} and FernSVM, a supervised Generative-Discriminative model. 300 positive samples and 900 negative samples are collected and labelled manually from S2 sequence to train the offline learning classifiers. In the Hx sequence of the Vehicle Dataset, the number of supervised positive and negative training data become 200 and 500, respectively.

\begin{figure}[!h]
\centering
\includegraphics[scale=0.5]{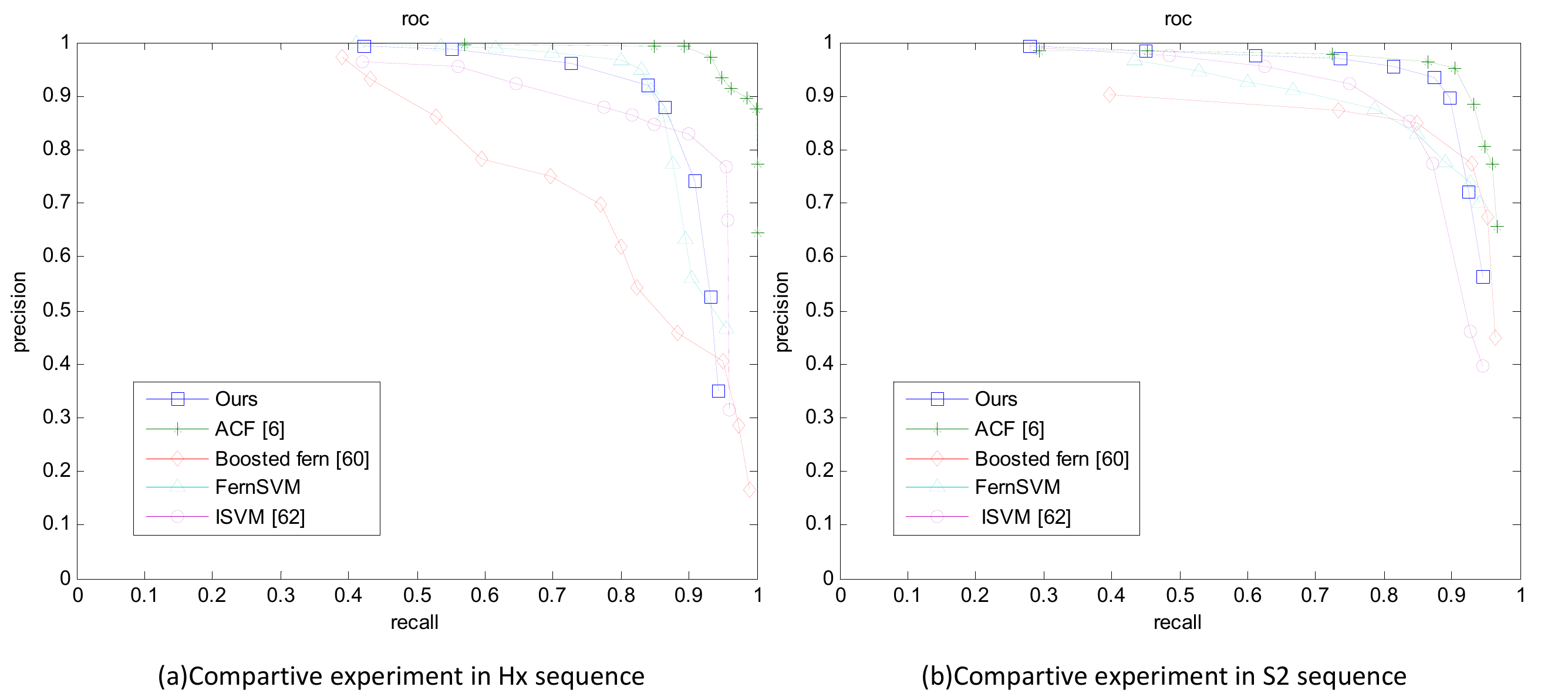}
\caption{Comparison with supervised learning methods}
\label{fig_9}
\end{figure}

\begin{figure}[!h]
\centering
\includegraphics[scale=0.5]{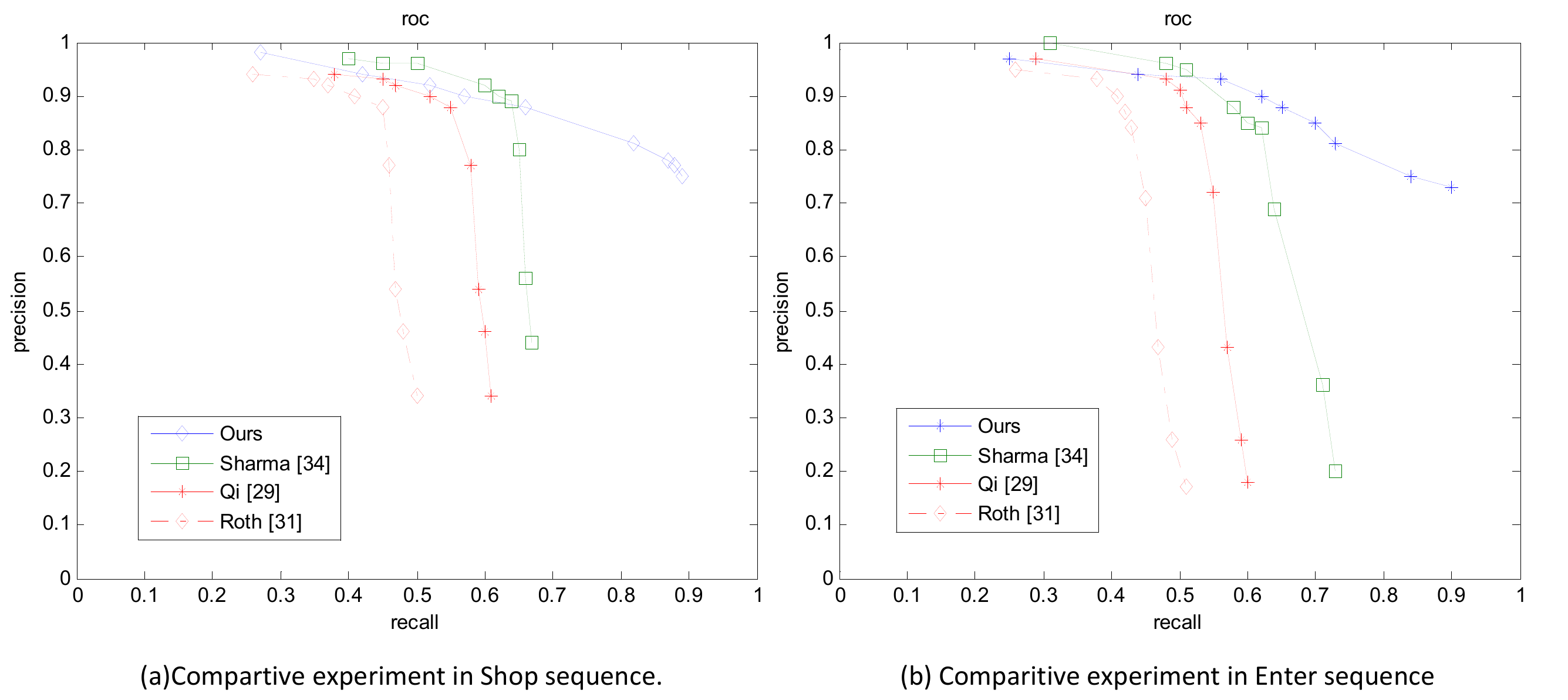}
\caption{Comparison with online learning methods}
\label{fig_10}
\end{figure}

\begin{figure}[!h]
\centering
\includegraphics[scale=0.7]{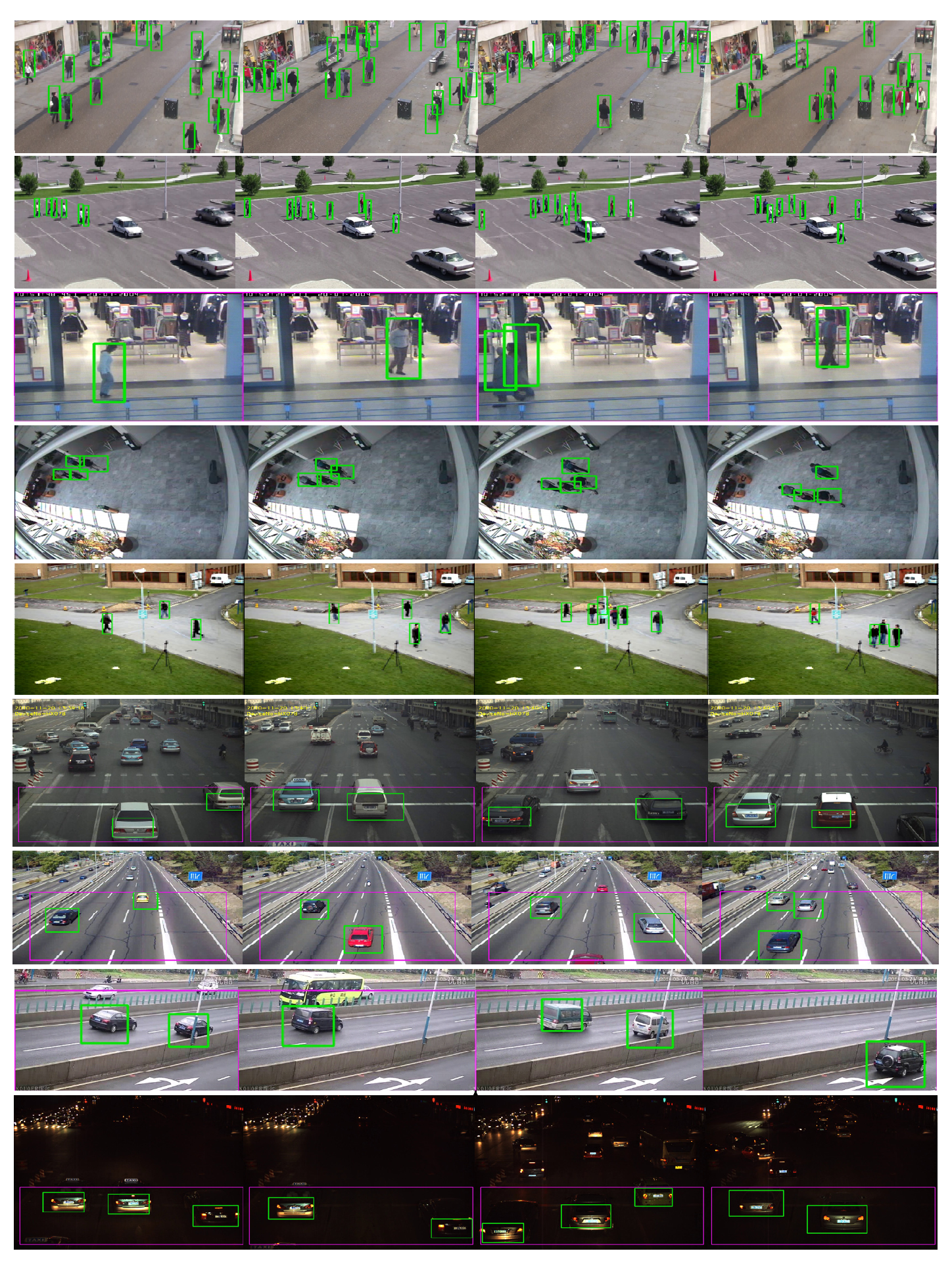}
\caption{Some detection results using our approach in Vehicle, GRAM-RTM, Towncenter, PNN-Parking-Lot2/Pizza, CAVIAR and PETS2009 Datasets. Note that the Hx, Yk, Hi and Hn sequences have high resolutions. Thus, a ROI region of purple box was settled for improving the detection speed. The detection results were shown in the last three rows.}
\label{fig_11}
\end{figure}

The ROC curves are shown in Fig. \ref{fig_9} and Fig. \ref{fig_10}. The according F-measures are calculated in Tab. \ref{tab4} and Tab. \ref{tab5}. The ACF method outperforms the proposed method on Hx and S2 sequences. Our approach outperforms the other online object detection methods and achieves detection performance competitive with the supervised methods. We demonstrate the primary reason for these results is our special strategy to address the hard samples, required to detect objects in cluttered environments. Detection results are shown in Fig. \ref{fig_11}.

\begin{table}[htbp]
\centering
\caption{COMPARISON WITH ONLINE LEARNING METHODS}
  \label{tab:ModelASymbol}
 \begin{tabular}{ccccc}
 \toprule
 Sequence & Roth[31] & Qi[29] & Sharma[34] & Ours \\
 \midrule
 Shop  & 0.5955 & 0.6769 & 0.7446 & 0.8225\\
 Enter & 0.5634 & 0.6454 & 0.6992 & 0.8061\\
 \bottomrule
 \end{tabular}
  \label{tab4}
\end{table}

\begin{table}[htbp]
\centering
\caption{COMPARISON WITH SUPERVISED LEARNING METHODS}
  \label{tab:ModelASymbol}
 \begin{tabular}{ccccccc}
 \toprule
\multirow{2}{*}{Method}&
    \multicolumn{2}{c}{S2}&\multicolumn{2}{c}{Hx}\\
    \cmidrule(lr){2-3} \cmidrule(lr){4-5}
     &F-Measure &FPS  &F-Measure &FPS \\
\midrule
Ours      &0.9036 &62 &0.8779 &10 \\
ACF[6]   &0.9271 &14 &0.9518 &5 \\
ISVM[62]      &0.8438 &2  &0.8624 &4 \\
Boosted fern[60]     &0.8496 &92 &0.7320 &11 \\
FernSVM  &0.8391 &62 &0.8771 &10 \\
 \bottomrule
 \end{tabular}
  \label{tab5}
\end{table}

\subsection{Multi-view Object Detection in Video Sequences}

\begin{figure}[!h]
\centering
\includegraphics[scale=0.50]{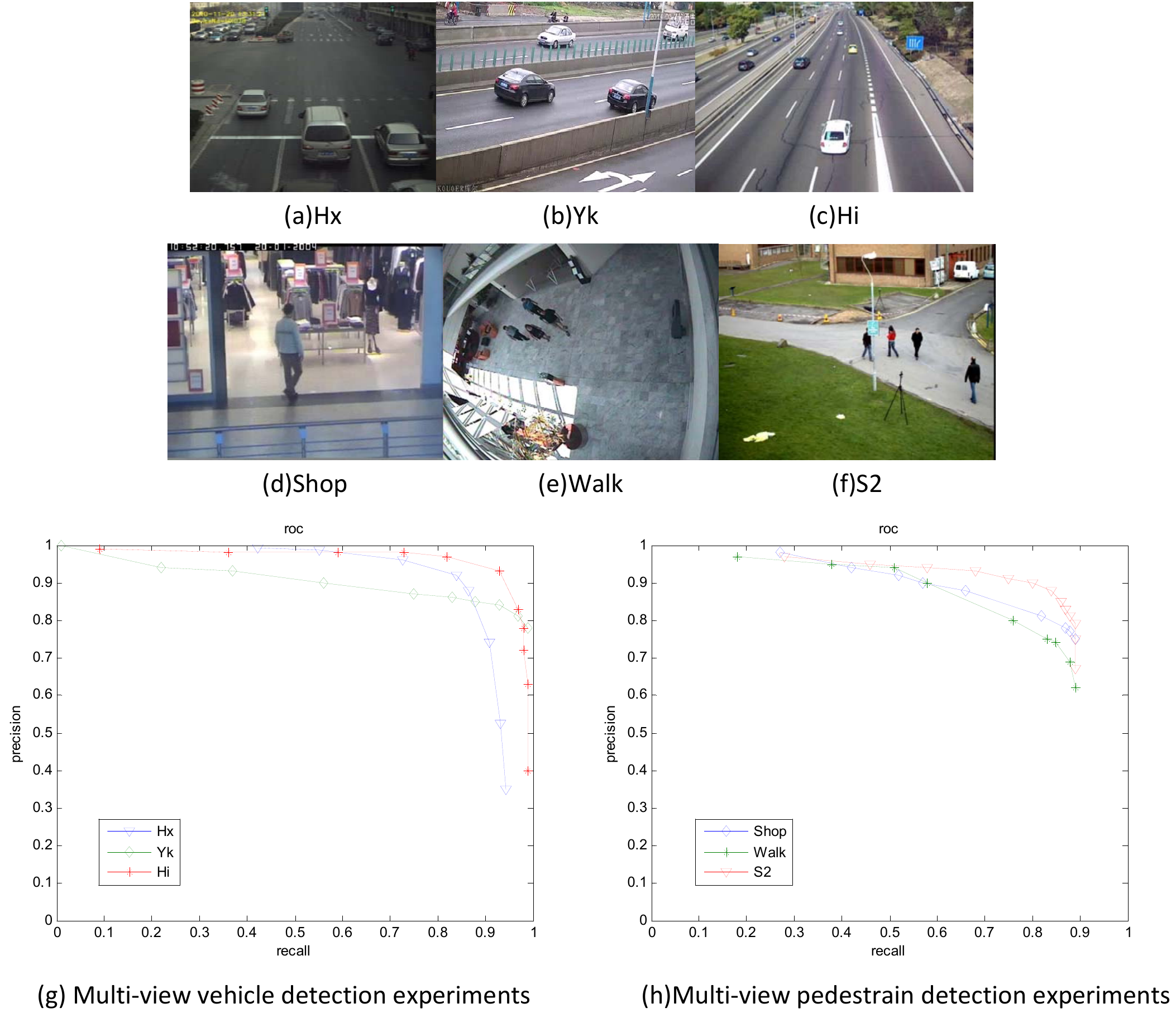}
\caption{Multi-view vehicle and pedestrian detection experiments.}
\label{fig_multiview}
\end{figure}

Further tests check if the system can self-adjust to view point changes. Our system, initialized from several bounding boxes, was applied to the video sequences Yk and Hi from Vehicle and GRAM-RTM datasets. For the Yk sequence, the detector self-learns 156 positive samples and 323 negative samples. For the Hi sequence, 244 positive samples and 598 negative samples were automatically collected and labelled to online train the video-specific detector. As shown in Fig. \ref{fig_multiview}, the new detector achieves state of the art performance without any human intervention. Following this action, our framework is used for multi-view pedestrian detection in CAVIAR and PETS2009 dataset.  The number of online learning samples, self-learning duration and detection speed, different in each video, as shown in Tab. \ref{tab3}. Thus, the trained scene-specific detector can detect object in real time on a standard PC (Intel Core i5 3.2 GHz with 4 GB RAM). The detection performance is shown in Fig. \ref{fig_multiview}. It is valuable to note all self-learning processes are fully unsupervised without any prior knowledge or constraint. Our method can be easily employed in other surveillance scenes and form a bottom-up multi-view object detection method.

\begin{table}[htbp]
\centering
\caption{DISCRIPTION OF ONLINE LEARNING PROCESS}
  \label{tab:ModelASymbol}
 \begin{tabular}{ccccc}
 \toprule
 \multirow{2}{*}{Sequence} & Positive  & Negative  & Duration & Detection\\
                          & samples   & samples   &times(s)  & speed(FPS)\\
 \midrule
   Yk & 156 & 323 & 815 & 36 \\
   Hi & 244 & 598 & 2751 & 15 \\
   Hx & 336 & 687 & 25196 & 10 \\
   Shop & 281 & 479 & 1121 & 19 \\
   Walk & 410 & 311 & 2728 & 34 \\
   S2 & 504 & 994 & 9787 & 62 \\
 \bottomrule
 \end{tabular}
 \label{tab3}
\end{table}

%\begin{table}
%\center
%\caption{DISCRIPTION OF ONLINE LEARNING PROCESS}
%\begin{tabular}{|c|c|c|c|c|}
%\hline
%\multirow{2}{*}{Sequence} & Positive  & Negative  & Duration & Detection\\
%                          & samples   & samples   &times(s)  & speed(FPS)\\
%\hline
%Yk & 156 & 323 & 815 & 19 \\
%\hline
%Hi & 244 & 598 & 2751 & 8 \\
%\hline
%Hx & 336 & 687 & 25196 & 10 \\
%\hline
%Shop & 281 & 479 & 1121 & 13 \\
%\hline
%Walk & 410 & 311 & 2728 & 15 \\
%\hline
%S2 & 504 & 994 & 9787 & 9 \\
%\hline
%\end{tabular}
%\end{table}

\section{Conclusions and Discussions}

This paper presents a self-learning video object detection framework. In this framework, a Generative-Discriminative model can be trained by online gradual optimized process without any human labelled samples or generic detectors. This framework can be easily employed in multiple surveillance scenarios and will result in scene-specific detectors more dedicated by a hierarchical optimized process to the problematic samples. Consequently, the Generative-Discriminative model puts more emphasis on the most distinctive parts of the object category, which leads to the reduction of the global classification error. This process simulates the autonomous learning of human. Experimental results show our approach achieves high accuracy on multi-scene vehicle and pedestrian detection tasks.

Future investigations will integrate the self-learning detector with an online learning tracker to form a scene-specific multiple object detection and tracking system which will simultaneously increase the performance of the detection and tracking.

\section*{Acknowledgment}
This work was supported by the National Natural Science Foundation of China (61302137, 61603357, 61271328 and 61603354), Wuhan huanghe Elite Project, Fundamental Research Funds for the Central Universities Young Teacher Promotion Program-Outstanding Youth Foundation, China University of Geosciences (Wuhan)(CUGL170210), Fundamental Research Funds for National University, China University of Geosciences (Wuhan) (1610491B06).

%% The Appendices part is started with the command \appendix;
%% appendix sections are then done as normal sections
%% \appendix

%% \section{}
%% \label{}

%% References
%%
%% Following citation commands can be used in the body text:
%% Usage of \cite is as follows:
%%   \cite{key}          ==>>  [#]
%%   \cite[chap. 2]{key} ==>>  [#, chap. 2]
%%   \citet{key}         ==>>  Author [#]

%% References with bibTeX database:

\bibliographystyle{elsarticle-num}
\bibliography{newcontrolpaperref,shoslifref}

%% Authors are advised to submit their bibtex database files. They are
%% requested to list a bibtex style file in the manuscript if they do
%% not want to use model1-num-names.bst.

%% References without bibTeX database:

% \begin{thebibliography}{00}

%% \bibitem must have the following form:
%%   \bibitem{key}...
%%

% \bibitem{}

% \end{thebibliography}

\end{document}